\documentclass[pdflatex, sn-aps]{sn-jnl}

\usepackage{longtable} % Include in the preamble for longtable support
\usepackage{subfigure}
\usepackage{graphicx}%
\usepackage{multirow}%
\usepackage{amsmath,amssymb,amsfonts}%
\usepackage{amsthm}%
\usepackage{mathrsfs}%
\usepackage[title]{appendix}%
\usepackage{xcolor}%
\usepackage{textcomp}%
\usepackage{manyfoot}%
\usepackage{booktabs}%
\usepackage{algorithm}%
\usepackage{algorithmicx}%
\usepackage{algpseudocode}%
\usepackage{listings}%nal)
\usepackage{tcolorbox}
\usepackage{caption}

\theoremstyle{thmstyleone}%
\theoremstyle{thmstyletwo}%

\theoremstyle{thmstylethree}%

\begin{document}

\title[LLMPR]{\emph{LLMPR}: A Novel LLM-Driven Transfer Learning based Petition Ranking Model}


\author[2]{\fnm{Avijit} \sur{Gayen}}\email{avijit.g@technoindiaeducation.com}

\author[3]{\fnm{Somyajit} \sur{Chakraborty}}\email{somyajitchakraborty@ucc.ie}

\author[2]{\fnm{Mainak} \sur{Sen}}\email{mainaksen.1988@gmail.com}

\author[2]{\fnm{Soham} \sur{Paul}}\email{soham211001001105@technoindiaeducation.com}
\author*[1]{\fnm{Angshuman} \sur{Jana}}\email{angshuman@iiitg.ac.in}

\affil*[1]{ \orgname{India Institute of Information Technology Ghuwahati}, \orgaddress{\street{Bongora}, \city{Ghuwahati}, \postcode{781015}, \state{Assam}, \country{India}}}

\affil[2]{ \orgname{Techno India University, West Bengal}, \orgaddress{\street{Salt Lake}, \city{Kolkata}, \postcode{700091}, \state{West Bengal}, \country{India}}}

\affil[3]{ \orgname{University College Cork}, \orgaddress{\street{Collage Rd.}, \city{Cork}, \postcode{T12 K8AF}, \state{Cork}, \country{Ireland}}}


\abstract{
The persistent accumulation of unresolved legal cases, especially within the Indian judiciary, significantly hampers the timely delivery of justice. Manual methods of prioritizing petitions are often prone to inefficiencies and subjective biases, further exacerbating delays. To address this issue, we propose \textbf{LLMPR} (Large Language Model-based Petition Ranking), an automated framework that utilizes transfer learning and machine learning to assign priority rankings to legal petitions based on their contextual urgency. Leveraging the ILDC dataset comprising 7,593 annotated petitions, we process unstructured legal text and extract features through various embedding techniques, including DistilBERT, LegalBERT, and MiniLM. These textual embeddings are combined with quantitative indicators such as gap days, rank scores, and word counts to train multiple machine learning models, including Random Forest, Decision Tree, XGBoost, LightGBM, and CatBoost. Our experiments demonstrate that Random Forest and Decision Tree models yield superior performance, with accuracy exceeding 99\% and a Spearman rank correlation of 0.99. Notably, models using only numerical features achieve nearly optimal ranking results (R\textsuperscript{2} = 0.988, $\rho$ = 0.998), while LLM-based embeddings offer only marginal gains. These findings suggest that automated petition ranking can effectively streamline judicial workflows, reduce case backlog, and improve fairness in legal prioritization.
}




\maketitle

\section{Introduction}\label{s:intro}
The judiciary system has been a part of any nation since ancient times. Historically, it was controlled and served by a monarch of a kingdom or an empire. Since then, the judiciary system has become a separate, independent part of any nation, irrespective of the nature of the government. It not only serves as a protector of the constitution of the nation but also ensures the fundamental rights of the citizens of the country. The function of the judiciary system is to provide timely justice to its citizens, irrespective of their caste, colour, and financial status. The smooth functioning of the judiciary system ensures the prosperity of the nation. The timely availability of legitimate justice enhances the morale of society, hence it reduces corruption as well as improves the social systems. On the other hand, the delayed process of the judiciary system deprives the common citizen of their deserved justice~\cite{melcarne2021justice}. \cite{joshi2023addressing}examines the factors like legal infrastructure, out-of-date laws which influence the delay in judiciary decisions.  

% \begin{figure}[h]
% \centering
% \includegraphics[width=0.43\textwidth]{samples/court.PNG}
% \caption{Indian Judiciary System Hierarchy}
% \label{fig:court}
% \end{figure}
\par The backlog in the judiciary system has been observed as one of the major issues. It is an open truth that delayed justice is always compromised due to lack of evidence which is abolished over time. In the Indian judiciary system, currently, around $30$ million cases are pending. According to statistics, approximately $73000$ cases are pending per judge~\cite{singh2018indian}. The existing complicated paperwork and rigid rules of assessing justice make it a more time-consuming process. Thus, the accessibility of justice for the common marginal people of society is far off. On the other hand, this delayed process of justice is capitalised on by the rich people for their own self-interest \cite{baruah2012judicial}. This situation develops partiality in the judicial system. Moreover, political and rich people disrupt the independence of the judiciary system \cite{kumar2023impact}. Also, the lack of coherent data across the different courts and such fixed deadline for the completion of cases makes this situation worse. Along with the large set of adjourned cases, the increasing crime rate over the years and increasing count of new Public interest litigation (PILs)\footnote{{PIL is type of legal action where an individual or group initiates a case to protect the rights and interests of the general public or a marginalized group, even if they haven't been directly harmed. Essentially, it's using the legal system to address issues of public concern and social change \cite{cummings2009public}.}} overload the judiciary system day-by-day~\cite{hazra2020does}. Finally, the grey area of the constitution majorly hinders the fast judgment process in those cases which deal with the fundamental rights of the citizens. 
\par There are fast-track courts and separate courts, tribunals, etc. that have been set to prioritise and quickly settle cases by expert judges. However, the introduction of technology became indispensable to handle backlogs, data coherency, etc, in the judiciary system. According to the Indian Law Commission~\cite {lawc}, it has been suggested to reduce the oral argument time in court proceedings by the introduction of technology-based proceedings to improve the backlogs. In this context, legal judgment prediction has become 
a point of interest for researchers in the last few years in the domain of machine learning and artificial intelligence. In ~\cite {lawlor1963computers}, the authors mainly focused on predicting judgment in the context of various cases. The ever-increasing new petitions overloads the judiciary system and becomes the main issue of the backlog of this system. This backlog not only overloads the entire judiciary system but also delays important cases due to the long list of comparatively trivial cases. The manual ranking of petitions often faces biases which hinder the urgency and fast decision of the petitions of sensitive and concurrent issues. Fast-track courts exist to address the high-priority cases; however, a quick and automated ranking system of the petitions can negate the manual biases. Introduction of such a system not only negates the manual biases but also identifies the importance of the petition to rank them, which in turn accelerates the entire judiciary system to reduce the existing backlogs. There are very few works~\cite{vercosa2024investigation,sokhansanj2022predicting,faccioni2023applying} that are related to the above issue, we could not find any such work that could address the issue. Thus, there is a strong requirement for the development of such a petition ranking system, which will reduce the backlogs by prioritising the petitions which help them to get a quick final decision on petitions. 
% \par \textbf{Motivation:} The major issues that the Indian judiciary system faces are 
% backlog of cases, shortage of judges in court, corruption, extra payment to make cases fast track \cite{singh2018indian}. Independence and an impartial judiciary system is necessary for any country. Indian democracy being such large, it is quite hard to maintain an impartial and timely decision-making process. Legal judgment prediction(LJP) is a crucial and time-consuming task that requires extensive legal documental research and fact collection \cite{lawlor1963computers}. Analysis of the collected facts with reference to previous cases needs a lot of time and effort. Currently, this task is carried out by legal experts. In India, currently, we have 1000 legal cases pending and 50 is the number of judges. For this reason, low-income people in India find it very difficult to get legal advice at a proper cost. The petitioners and respondent both pays huge money and time to get justice. Citizens need to wait for days and months to get the date of hearing as xxxx number of cases is allocated to yyyy of judges.
\par In this work, we predict the rank of  petitions' which has been initially accepted for further legal proceedings. Specifically, we predict the rank of accepted petitions filed based on the statement of the petition framed by the legal practitioner. This ranking system would work as an automated system that identifies the importance based on contextual sensitivity and fundamental judicial priority of the petitions filed, but also helps to automatically negate the pressure of trivial cases, cases with the least legal basis, etc. Hence, precious court time is being saved and allowing discussion of more important cases that are relevant to social improvements. The major challenges of this work could be 
\begin{itemize}
    \item petitions are unstructured, i.e., these are different from one case to another.
    \item The languages could be different
    \item There exists a large corpus of data
\end{itemize}
Therefore, predicting the rank of initially accepted petitions is challenging as well as indispensable to developing a decision support system in this context. This system would not only reduce the backlog of the judiciary system but also provide a fair and unbiased judiciary system.

\par To predict the ranking of accepted petitions, we proposed the following methodology that leverages a structured framework using machine learning techniques. In our work, we have used the ILDC dataset~\cite{maliketal2021ildc}, containing $7,593$ annotated petition records, serves as the foundational data source. Initial preprocessing of the textual data focuses on handling unstructured text by eliminating noise such as stop words, special characters, and non-alphanumeric tokens. Techniques like stemming and lemmatisation ensure the data is standardized and simplified for downstream analysis. Further, we employ various advanced text embedding methods, including DistilBERT, LegalBERT, and MiniLM, to transform the text into meaningful numerical representations. Alongside these embeddings, we also use numerical features $gaps\_days$, i.e., time gap between acceptance of the petition to its first proceeding and word counts. These features are combined into a unified feature matrix to effectively train these machine learning models.
\par We further incorporate a wide range of machine learning models like Random Forest, Decision Tree, XGBoost, LightGBM, CatBoost, ElasticNet, and Linear Regression. These models are selected to capture complex relationships within the data. The ranking system uses a combination of numerical and textual features, where models predict the petition rank based on semantic relevance and procedural urgency. Evaluation metrics such as accuracy, precision, recall, F1-score, and Spearman rank correlation ensure robust assessment of model performance. Cross-validation techniques, including K-Fold and Monte Carlo, validate the generalizability of the models. Random Forest and Decision Tree models demonstrate superior performance, achieving high accuracy and rank correlation values, making them the most effective for petition ranking tasks. This systematic approach not only identifies high-priority cases but also contributes to reducing the judicial backlog by streamlining the ranking of petitions.
% \par {To develop a decision support system that would be able to predict the acceptance or rejection of the petition at the very initial stage, we have used supervised machine learning algorithms. In our work, we have used the ILDC dataset\cite{maliketal2021ildc} from which we have used around $7500$ data points. It has mainly two important columns, i.e., text which contains the verbatim records of each petition and another column containing the nature of the initial decision of the petition based on which it has been further proceeded in court. We specifically considered the problem as a binary classification problem that learns from the raw text of the filed petition and finally classified it as `accepted' or `rejected'. We have used tf-idf as the feature extractor for the text corpus. After obtaining the feature matrix, we used supervised machine learning algorithms namely logistic regression, support vector machine(SVM), random forest (RF), decision tree(DT), and naive Bayes (NB) to train whether a given petition can be accepted for further proceedings of the court or not. Our methodology showed that random forest provided the best accuracy of 79\% with 10\% of test size.} 
% \textcolor{red}{Add the new methodology adopted as well as the changes in the objective of the work along with the brief results of the experiments.}
\par The novelty of the work is to predict the rank of the ``accepted"  petitions using machine learning techniques from the unstructured legal petition data. The contribution of the work can be listed as follows---
\begin{itemize}
    \item We have employed various LLM-based state-of-art text embedding techniques including domain specific text embedding method i.e., LegalBERT to convert the text into feature vector.
    \item We have also included various numerical features, i.e., $gap\_day$, $rank\_soore$, $word\_count$ and $sentence\_count$ in our model. 
    % \item We have applied machine learning algorithms to predict the initial decision of acceptance of the legal petition.
    % \item We have developed various features i.e., tf-idf, word2vec and countvectorizer from the raw petition text.
    \item We have measured the performance of several machine learning algorithms on our dataset to identify the most suitable model in this context.
    \item Finally, we have validated our model prediction with the actual labels of each petition against our ranking obtained by our proposed method using spearman rank correlation. We found Random Forest gave the best result. 
\end{itemize}

{Table \ref{tab:abbreviations} presents all the abbreviations and terminologies that we shall use in the course of this study. In the next section, we highlight the major related works.} In section ~\ref{s:proposed}, we discuss the proposed work for the petition ranking framework, and further results are shown in section~\ref{s:result}. We finally conclude with the future direction of this work in section~\ref{s:conclusion}.

\begin{table}[h!]
\centering
\caption{{List of Abbreviations}}
\begin{tabular}{|l|p{8cm}|}
\hline
{\textbf{Abbreviation}} & {\textbf{Definition}} \\
\hline
{LLM} & {Large Language Model} \\
\hline
{LLMPR} & {Large Language Model-based Petition Ranking} \\
\hline
{PIL} & {Public Interest Litigation} \\
\hline
{BERT} & {Bidirectional Encoder Representations from Transformers} \\
\hline
{TF-IDF} & {Term Frequency-Inverse Document Frequency} \\
\hline
{RF} & {Random Forest} \\
\hline
{XGBoost} & {Extreme Gradient Boosting} \\
\hline
{LGBM} & {Light Gradient Boosting Machine} \\
\hline
{CatBoost} & {Categorical Boosting} \\
\hline
{ELNet} & {ElasticNet Regression} \\
\hline
{LR} & {Linear Regression} \\
\hline
{NN} & {Neural Network} \\
\hline
{ILDC} & {Indian Legal Documents Corpus} \\
\hline
{MSE} & {Mean Squared Error} \\
\hline
{MAE} & {Mean Absolute Error} \\
\hline
{Sp. Corr.} & {Spearman Rank Correlation} \\
\hline
{Exp. Var.} & {Explained Variance Score} \\
\hline
\end{tabular}
\label{tab:abbreviations}
\end{table}

% \par \textbf{Contribution \& Paper organization}
% The major contribution of the paper are as follows---1. We have used machine learning algorithms to automate the decision-making process. 2. We have compared the results of different machine learning algorithms and seen that random forest provides better results on our dataset. After this brief introduction in section \ref{introduction}, in section \ref{survey}, we have presented the literature survey. Next, our proposed methodology is presented in \ref{method}. Section \ref{result} deals with the experimental results and finally we conclude with the future scope in section \ref{scope}.

\section{Related Work}\label{s:related}In this section, we discuss a comprehensive survey of the related works aligned to the proposed work. Initially, we highlight the works related to the judiciary systems and their issues. Further, we discuss some recent relevant works of machine learning techniques that have been used to mitigate the litigation of the judiciary system. We also outline some recent works on petition decision support system. Finally, we highlight the works related to AI-enabled judiciary framework in recent times. These works find the research gap in current trends of works in the context of AI in judiciary system.
\subsection{Judiciary System \& Litigation of the system} The Judiciary system has been introduced as one of the parts of the social system from the ancient age of civilization. Based on the belief, social, and religious values, the judiciary system of any nation pertinently includes those in their legal justice~\cite{abdillah2023indonesian,kholiq2023does}. Though the legal proceedings of any nation develop based on social and religious values, it has a strong impact and influence across the nations due to the exchange of culture, education, trade, etc. Though the smooth functioning of the judiciary system ensures timely justice and hence improves the legal rights of citizens, the judiciary system faces a lot of potential issues irrespective of the country~\cite{sil2023review,ash2022measuring, ippoliti2020efficiency}.  
\par The delay in delivering justice has been a major issue of the judiciary system\cite{melcarne2021justice,singh2018indian}. In \cite{singh2018indian}, Singh et al. observed that the average period of delivery of justice is over multiple decades in the Indian judiciary system. 
% The backlog in the judiciary system has been observed as one of the major issues. It is an open truth that delayed justice is always compromised due to lack of evidence which is abolished over time. 
According to the author, in the Indian judiciary system, currently approximately $30$ million cases are pending where around $73000$ cases are pending per judge. 
% The existing complicated paperwork and rigid rules of accessing justice make it a more time-consuming process. 
In some other works~\cite{sundari2021weakness, susanto2020court}, authors describe the hindrance of accessibility of justice by the common marginal citizens and how corruption has influenced the judicial system. It also discusses the suppression of legitimate justice under the strong control of political influence. 
% On the other hand, this delayed process of justice is capitalized on by the rich people for their own sake of interest. This situation develops partiality in the judiciary system. Moreover, political and rich people's influence disrupts the independence of the judiciary system~\cite{sundari2021weakness,susanto2020court}. 
In some works~\cite{vcehulic2021perspectives}, authors highlighted the lack of data coherency across the different courts. In another work~\cite{barno2020exploring}, Barno et al. observed that the absence of a proper schedule and deadline for the completion of cases makes the system inefficient. In some work~\cite{hazra2020does}, the author pointed out that the increasing crime rate over the years as well as the increasing count of new Public interest litigation (PILs), overload the judiciary system day by day. Along with all the above issues, the lack of interpretation of the fundamental rights of the citizens of the constitution majorly delays the judgment process~\cite{farrell2007excess}. 

\subsection{Machine Learning in Judiciary System} While fast-track courts and specialized tribunals have been established to expedite case resolution through expert judges, the integration of technology has become essential in addressing judicial backlogs and ensuring data coherence in the legal system. In a recent work~\cite{chawla2022commerce}, authors observed that strong enactment of consumer protection law would improve the growth of e-commerce business. As per the recommendations of the Indian Law Commission, it has been proposed to decrease the duration of oral arguments in court proceedings~\cite{sourdin2020court} by incorporating technology-based proceedings as a means to alleviate backlog issues~\cite{rasheed2021alternative,smith2020integrating}. In this context, the prediction of legal judgments has garnered attention from researchers in recent years within the field of machine learning and artificial intelligence\cite{zhong2020does}. These works~\cite{lawlor1963computers} mainly focus on predicting judgment in the context of various cases. Some relevant works~\cite{shi2021smart,putra2020modern} discussed the introduction of e-court to enhance the judiciary system. ~\cite{benedetto2025leveraging}used explodes the use of LLM(Large Language Model) for summarizing Italian legal news. They showed that LLM outperforms older models like BART, T5 in terms of grammatical accuracy. Authors in ~\cite{wang2024causality}, experimented on the CAIL2018 dataset to show superior accuracy and robustness. In this work, the author combined semantic matching with causal relationship learning.  
\subsection{Petition Decision Support System}
The continuous influx of new petitions overwhelms the judicial system, emerging as the primary cause of its backlog. Consequently, there is a pressing need for a rapid and automated decision support system to handle the initial determination of petitions before proceeding to court discussions. While there are limited studies~\cite{li2017modeling} tackling the mentioned issue. In this work, the author discusses the method of predicting the initial decision of a petition filed in a web portal. Though the work deals with a structured dataset, as the data is collected from a web portal, there is a significant demand for the creation of a decision support system to decrease the volume of petitions at the initial stage. This, in turn, would contribute to the reduction of the backlog in the judicial system.
The authors in~\cite{sun2022autonomous}, 
proposes a hybrid deep learning model based on CNN and Bi-LSTM to automatically extract features from the title and body of a petition. Authors ~\cite{buryakov2024multi} have proposed a BERT-based fine-tuned multi-label classifier on a dataset from the Taiwanese Joint Platform. In ~\cite{ahmad2022hybrid}, a decision support system was made using CNN+BiLSTM to predict the court decision based on past data. Zekun et al. in ~\cite{yang2023explainable} proposed an explainable convolutional neural network model to enhance the e-petitions tagging system on the Message Board for Leaders (MBL) in China. This system uses layerwise relevance propagation (LRP), an understandable and explainable method to find 
interpretability compared to several baseline models.

\subsection{AI-enabled Judiciary Framework}
% \textcolor{red}{ Add the related work for the newly added part of the method---Argument Mining, transfer Learning, Explainable AI, LLM related works etc.}

LawGPT\_{zh} is a Chinese language model based on ChatGLM-6B LoRA 16-bit instruction. It contains legal question-and-answer datasets. LaWGPT ~\cite{nguyen2023brief}, a series of models pre-trained on large scale chinese legal text databases. It is also fine tuned on legal dialogue question-and-answer datasets. Lawyer LLaMA ~\cite{huang2023lawyer}, a Chinese legal LLM provides legal advice and generates legal advice. ~\cite{clavie2021legalmfit}showed that lightweight LSTM language models achieved better results on short legal text classification with reduced computational overhead compared to larger models. % ~\cite{janatian2023text} 
\par In the recent trend, AI and Law become a pivotal point of interest of the researchers. It produces a plethora of works that help to accelerate the slow judiciary system. In spite of the ample contribution in recent work in this domain, we do not find any contributory work that address the ranking framework for the ``Accepted" petitions that can not only helps in reduction of existing backlogs but also biases of manual intervention. Thus, there is a strong requirement of development of such petition ranking framework to accelerate the slow judiciary system. 

% \par Although we have seen that over the last few years many researchers along the world have tried to apply learning based algorithms to speed up the judiciary decision-making process but still we have not found any work that ranks the petitions according to their importance. Here, we have developed a machine learning-based system that takes a petition as input and finds its ranking.

\section{Proposed Work} \label{s:proposed}
This section describes the proposed model for the ranking of petitions. We initially describe the details of the legal labeled corpus used in this work. We further describe the methodology adopted to rank the "Accepted" petition. We also outline the features used in the proposed model. It includes the data pre-possessing method used to clean the dataset and the various text embedding techniques used in our work. We further describe the various machine learning model adopted in our proposed method.  %  We also provide a schematic diagram in figure~\ref{fig:schematic} to show the workflow to execute the method.

\subsection{Dataset}In this work, we use ``ILDC" (Indian Legal Documents Corpus) dataset~\cite{maliketal2021ildc}. It is a large corpus of over $35000$ Indian Supreme Court cases that has been annotated with original court decisions. A subset of this large corpus is used as a dataset that contains $7593$ data points. and In the dataset, we have 4 columns: a)	\textbf{`text’}, b)	\textbf{`label’}, c)	\textbf{`split’}, d) \textbf{`name’}. The description of the columns of the dataset is as follows:
\begin{itemize}
    \item{\bf Text}: The ‘text’ column actually consists of the annotated cases in plain text. The data-type of this column is ‘object’. There are correspondingly 7593 text recordings as there are no ‘Null’ values in this column. Each data point, i.e., petition/case-recording has an average length of 20000+ words (that means, each case recording has average 20000+ words).
    \item{\bf Label}: The ‘label’ column actually holds the initial decision of the corresponding petition/case-recording. It has values of 1 and 0, which signifies what decision was granted – whether the petition was accepted (1) or rejected (0). This column has a data-type of ‘int64’ and also has 7593 values, without any empty/null recordings. There are correspondingly 3194 cases which were accepted (decision – 1) and 4399 cases which were rejected (decision – 0).
    \item {\bf Split}: The ‘split’ column has a data-type again of ‘object’ and basically helps us identify the splitting of the data. This column has 3 types of values – Train, Test and Development. The number of petitions belonging to these 3 categories are as follows: Training has 5082 petitions assigned, Testing has 1517 petitions assigned, and Development has 994 assigned to it. This signifies that 5082 petitions (texts) would be used for training the model, 1517 petitions would be used for testing and 994 petitions would be used for development.
    \item{\bf Name}: The last column also has ‘object’ as it’s data type. This column basically tells us the name of the ‘.txt’ file where the case-recording/petition had been taken from. The filename generally has the format $‘year\_caseno.txt’$, thus the names of the files in this column have been entered like that, first the year of the filing, then the case number, like for example, $‘2008\_1460.txt’$
\end{itemize}

{To avoid data leakage, we respected the predefined \texttt{split} column in the ILDC dataset. Train and test partitions are non-overlapping. We further confirmed this by computing pairwise TF--IDF cosine similarity of bigrams across splits, which yielded a maximum similarity of 0.765---well below the 0.80 threshold typically used to flag near-duplicates.}
{To generate a continuous urgency ranking for each petition, we computed the number of days between the petition filing date and its first listed court hearing (denoted as \textit{gap\_days}). This value was then transformed using inverse square scaling and log normalization to obtain the final \textit{rank\_score}, where smaller delays result in higher urgency. This method is inspired by prior judicial delay studies that treat early court attention as a proxy for systemic priority.}

% We have used the ILDC(Indian Leagal Document Corpus) single dataset{\cite{maliketal2021ildc} for this work. It is a comprehensive collection with a total of $7,593$ instances, thoughtfully divided into three distinct subsets for various purposes within natural language processing (NLP) and machine learning tasks. Here are additional details about the dataset: `text' contains the pre-processed data. `label' contains either `0' or `1'. `0' represents the rejection of the concerned petition and `1' represents that the concerned petition has been accepted. The `split' column maintains that the file belongs to either train set, validation set, or test set. The `name' field shows the name of the file which stores the additional information regarding the petition. 
The snippet of the dataset has been shown in the table~\ref{table:dataset_snippet_Paper}.
\begin{table}[h!]
 \centering
  \begin{tabular}{l|l|l|l}
  \hline
    Text & Label & Split &Name\\
    \hline
    F. NARIMAN, J. Leave granted. In 2008, the Pu... & 1 & train & 2019\_890.txt\\
   S. THAKUR, J. Leave granted. These appeals ar... & 0 & train & 2014\_170.txt\\
    Markandey Katju, J. Leave granted. Heard lear... & 1 & train & 2010\_721.txt\\
    civil appellate jurisdiction civil appeal numb... & 0 & dev & 1989\_75.txt\\
     original jurisdiction writ petitions number. 8... & 0 & dev & 1985\_233.txt\\
     civil appellate jurisdiction civil appeal numb... & 1 & test & 1986\_397.txt\\
     criminal appellate jurisdiction criminal appea... & 0 & test & 1993\_98.txt\\\hline
  \end{tabular}
  \caption{Format of ILDC dataset}
  \label{table:dataset_snippet_Paper}
\end{table}
\subsection{Methodology}
\begin{figure}[h]
\centering
\includegraphics[width=1\textwidth, height=1.5 in]{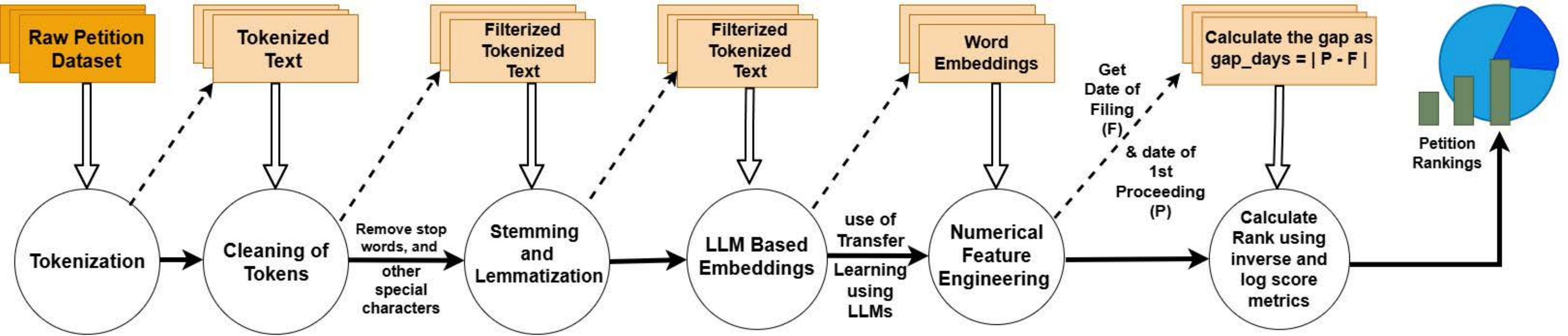}
\caption{Schematic diagram of Petition Ranking Model using Machine Learning. }
\label{fig:schematic}
\end{figure}

In this section, we have described a detailed methodology adopted in our proposed model. The methodology adopted in this study integrates robust preprocessing techniques, most relevant text embedding techniques, state-of-the-art machine learning algorithms, and evaluation metrics to develop a predictive framework for legal petition rankings. The framework combines numerical features extracted from the petition using LLM-based technique and textual embeddings extracted from the large size text petition. In this section, we outline the theoretical foundations, mathematical formulations, and practical implementations of the proposed model. The methodology involves the following key steps:
\begin{enumerate}
    \item Data Preprocessing: As the petitions are majorly unstructured in nature, Initially we cleaned and preprocessed it to remove noise and irrelevant information. This included steps such as removing stop words, tokenization, stemming and lemmatization.
    \item Text Embedding: In this work, we further use various LLM-based text embedding techniques  to convert the textual data into numerical features suitable for machine learning algorithms.
    \item Numerical Feature Integration: We next incorporate some derived numerical feature extracted from the text of the each petitions except the text embedded features with the help of OpenAI's GPT4o prompts \cite{hurst2024gpt}. The integration of this feature improves the rank predictability of our model.
    \item Ground Truth Preparation: We further prepare rank of the petitions from the extracted data from the text of the petitions. The ranking based on the extracted numerical score from the text of the petition is used as the ground truth of the proposed model for validation the testing.
    \item Model Selection \& Training: We use seven most relevant ML model e.g., Random Forest, Linear Regression, ElasticNet, Decision Tree, XGBoost, LightGBM,CatBoost. for this petition ranking model. These models was chosen for its balance between interpretability and performance in this context.
    \item Model Evaluation: At the end of the work, the model's performance was evaluated using metrics such as accuracy, precision, recall, and F1-score and AUC. The comparison based on the above metric reveals its ability to correctly rank the ``Accepted" legal petitions.
    % \item Explainability with LIME: To ensure transparency and interpretability, LIME (Local Interpretable Model-agnostic Explanations) was used to explain the model's predictions. LIME helps in understanding which features (words or phrases) contributed to a particular prediction, making the model's decision-making process more transparent. 
\end{enumerate}
We also include the schematic diagram in fig~\ref{fig:schematic} which describe the detailed flow of the work adopted in the proposed technique to predict the desired rank of the ``Accepted" petitions for further judicial processing after its initial decision. The raw petition dataset (ILDC Dataset) is taken, and with the help of Natural Language Processing (NLP), each petition is converted into tokens (tokenization), i.e., the words are constructed and represented as tokens for better representation and use for the model. Next, the tokenized texts or tokens are cleansed again with the help of NLP, where all stop words (words which are not of much value, like ‘a’, ‘an’, ‘the’, ‘he’, etc.) along with punctuation marks, alphanumeric characters and special characters. Once cleaned, the newly filtered tokens are passed through the processes of ‘Stemming’ and ‘Lemmatisation’, which basically reduce words to their root forms or dictionary forms based upon context, for better understanding and simplification. From these fully preprocessed tokens, word embeddings are generated with the help of LLMs. GPT-based LLMs are used so that word embeddings are specific, meaningful and contextual. Again, GPT-based LLMs help with the feature engineering part, from where we sort out ‘Date of Filing of the petition/case’ and ‘Date of the first hearing/proceeding’, for each petition. Once we get these 2 dates, it is easy to calculate the gap between these days. So, we calculate the gaps between the dates for each petition. Finally, we use metrics like inverse score and log score for ranking these petitions according to the urgency of their dates. The results are taken out using several models, and hence, the results from all the models are evaluated by checking the correlation between the results produced from each of them. This is achieved by Spearman Rank Correlation, which finally tells us which model’s ranking is preferred.
% \textcolor{red}{Describe the flowchart in details.---At first, the petitions are collected and then tf-idf feature extraction method was applied. The collected vectors are then used to train machine learning algorithms.In the detection phase, we enter new petitions to predict the outcome in the form either as accepted or rejected. The complete flow diagram is shown in figure~\ref{fig:schematic}.}

 \subsection{Features used in Learning model}In this section, we describe the features used for the classification and the pre-processing method applied before performing LLM-based text embedding vectorization techniques of the text field of the dataset to deduce the feature vectors. We further describe the method of numerical feature engineering which is being integrated in our model.
 % We have used citation-based as well as non-citation-based features in our model. In table~\ref{tab:ML_Features}, we list the features used and given a brief description of the each feature. 
 % As the dataset contains the labels of each data points, we have used supervised learning to develop prediction model. 
 % Though the citation-based features available in the dataset do not require any preprocessing before feeding the data into classifier, the non-citation-based features except publication count require the preprocessing.

  \subsubsection{Pre-processing of Large text Petition} We initially perform rigorous text preprocessing of the large size petitions( Average number of words/ petition $\thickapprox 30,000$). At the very first step, we filter the raw petitions from the dataset which are labeled as ``Accepted" after the initial decision of the petition. To handle the raw unstructured text data, we perform tokenization on the text field to create a structured input for further analysis. Next, we eliminate stop words that do not contribute much to the meaning of the text. Further, we remove non-alphanumeric characters, symbols, and punctuation marks from the tokenized text. This step helps to focus on the essential words and maintain consistency. We finally apply stemming and lemmatization techniques to reduce words to their base or root words. We also identified and handled rare words, either by removing them or replacing them with a common token. This technique has been adopted to prevent the model from overfitting on infrequently occurring terms. The major challenges faced to preprocess the textual data of the legal petitions are its voluminous in size as well as unstructured format. This preprocessing not only involves in standardizing the data to ensure uniformity but also eliminate noise to enhance the quality of downstream analysis. The preprocessing pipeline begins with converting all text to lowercase, which eliminates discrepancies due to case sensitivity. Non-alphanumeric characters, such as special symbols and punctuation, are removed to retain only meaningful tokens. 

% Lemmatization, a linguistic technique, reduces words to their root forms, ensuring semantic consistency. For instance, variations like "judging," "judged," and "judge" are reduced to their base form, "judge." Mathematically, this is expressed as:
%     \begin{equation}
%     \text{Lemma}(w_i) = w_{\text{root}},
%     \end{equation}
% where $w_i$ is the input word, and $w_{\text{root}}$ is the root form. Tokenization then splits sentences into individual tokens, enabling granular analysis of the text. For example, the sentence "The court accepted the petition" is tokenized into \{"The," "court," "accepted," "the," "petition"\}.

\subsubsection{Large Language Model Based Embedding} In this section, we discuss the detailed mathematical model and its understanding of the various text embedding techniques employed in our proposed model. We employed a suite of state-of-the-art Large Language Models (LLMs), each tailored to capture unique aspects of semantic understanding and contextual relationships in legal texts.

\paragraph{\bf \emph{DistilBERT}} This embedding technique is a distilled version of BERT embedding method. It was utilized for its efficiency and scalability. DistilBERT \cite{DBLP:journals/corr/abs-1910-01108} retains 97\% of BERT's language understanding capabilities with only 60\% of the parameters. Thus it makes suitable for resource-constrained environments. The embedding extraction process involves the transformer architecture's attention mechanism as introduced by Vaswani et al. \cite{vaswani2017attention}. The mathematical model used in \emph{DistilBERT} can be expressed as follows:
\begin{equation}
\mathbf{H} = \text{softmax}\left(\frac{QK^\top}{\sqrt{d_k}}\right)V,
\end{equation}
where $Q$, $K$, and $V$ are the query, key, and value matrices. The embeddings generated by DistilBERT ensure a balance between computational efficiency and semantic richness.

\paragraph{\bf \emph{MiniLM}}This text embedding technique is designed for lightweight applications. It employs deep self-attention distillation to achieve compact yet high-quality embeddings \cite{DBLP:journals/corr/abs-2002-10957}. Its mechanism focuses on preserving alignment between teacher and student model attention outputs, with the objective. The formal description of the model could be defined as follows:
\begin{equation}
\mathcal{L}_{\text{distill}} = \frac{1}{T} \sum_{t=1}^T \| \mathbf{A}_{\text{teacher}}^{(t)} - \mathbf{A}_{\text{student}}^{(t)} \|^2,
\end{equation}
where $\mathbf{A}$ represents the attention matrix. MiniLM demonstrated strong performance on recall tasks while maintaining computational efficiency.

\paragraph{\bf \emph{Flan-T5}} This embedding technique leverages pre-training objectives, including span corruption and multitask fine-tuning. It generalize across diverse tasks and models input-output pairs effectively through its encoder-decoder structure \cite{chung2022scalinginstructionfinetunedlanguagemodels}. With a given sequence $X$, it predicts output $Y$ by maximizing:
\begin{equation}
P(Y|X) = \prod_{i=1}^m P(y_i|X, y_1, \ldots, y_{i-1}),
\end{equation}
where $y_i$ represents each token in the output sequence. Flan-T5 proved particularly adept at handling complex relationships in legal texts.

% \paragraph{\bf \emph{Instructor-XL}} It is a multi-task embedding model, optimizes embeddings for diverse tasks simultaneously. Its architecture incorporates a task-specific guidance vector $\mathbf{g}$, with embeddings computed as:
% \begin{equation}
% \mathbf{E} = f_{\text{task}}(X, \mathbf{g}),
% \end{equation}
% where $f_{\text{task}}$ is a task-optimized transformation function. This approach ensures robust and adaptable embeddings suitable for multi-domain applications, including legal petition ranking.

\paragraph{\bf \emph{LegalBERT}} This text embedding method is a domain-specific variant of BERT. It was pre-trained on legal texts to capture the unique linguistic patterns and domain specific jargon \cite{DBLP:journals/corr/abs-2010-02559}. Its embeddings were fine-tuned to identify semantic and procedural nuances critical for legal analysis. The architecture follows the standard transformer setup, ensuring domain-specific relevance.

\paragraph{\bf \emph{E5}}This text embedding method is designed for general-purpose search and retrieval tasks and emphasize efficiency in semantic matching \cite{wang2024textembeddingsweaklysupervisedcontrastive}. The embedding process leverages contrastive learning objectives to ensure high-quality representations for similarity tasks. The formal representation of the technique can be stated as follows:
\begin{equation}
\mathcal{L}_{\text{contrastive}} = -\log \frac{\exp(\text{sim}(e_i, e_j))}{\sum_{k} \exp(\text{sim}(e_i, e_k))},
\end{equation}
where $\text{sim}(e_i, e_j)$ represents the cosine similarity.

% \paragraph{\bf \emph{RoBERTa}} It is an optimized version of BERT, utilizes dynamic masking and a larger pre-training corpus to enhance contextual understanding. Its embeddings, generated via a robust transformer architecture, excel at capturing semantic subtleties in lengthy petitions.

\subsubsection{Numerical Feature Engineering}To improve the raking predictability of the proposed model, we integrate few numerical features. This feature integration complements the textual embeddings by capturing temporal and structural aspects of the legal proceedings. We include following key features in our model: a) \texttt{gap\_days} i.e., the number of days between petition acceptance and the first proceeding, b)\texttt{rank\_score}, c) \texttt{word\_count}, and d) \texttt{sentence\_count}. The \texttt{gap\_days} feature is computed as:
\begin{equation}
\text{gap\_days} = |\text{date}_{\text{proceeding}} - \text{date}_{\text{acceptance}}|,
\end{equation}
It provides a direct measure of procedural delays in the judiciary system. We represent the two different scaling \texttt{rank\_score} i.e., a) \texttt{rank\_score\_log} and b)\texttt{rank\_score\_inverse\_square} to capture the different aspect of this score. The mathematical model of these features can be stated as:
\begin{equation}
\texttt{rank\_score\_log} = \log(1 + \text{gap\_days}),
\end{equation}
\begin{equation}
\texttt{rank\_score\_inverse\_square} = \frac{1}{\text{gap\_days}^2}.
\end{equation}
These transformations mitigate the influence of extreme values, improving interpretability and model robustness.

\begin{figure}[ht]
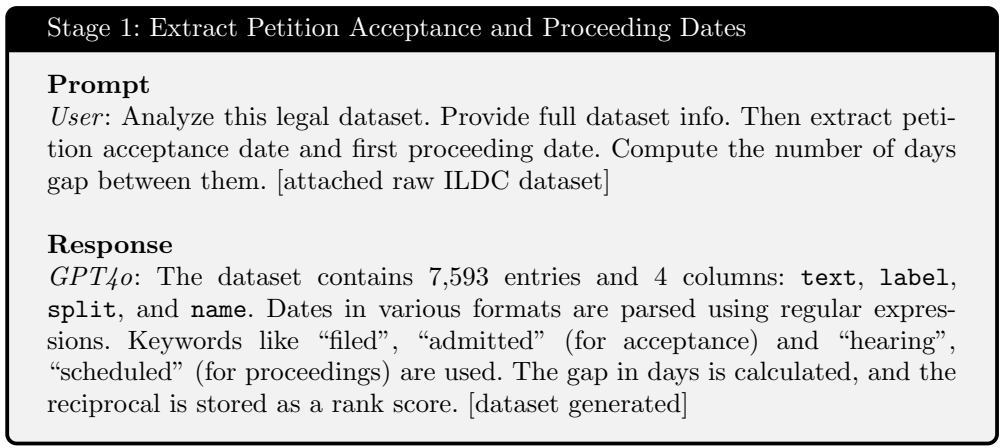

\centering
\begin{tcolorbox}[title=Stage 1: Extract Petition Acceptance and Proceeding Dates, colframe=black, colback=gray!10!white]
\textbf{Prompt} \\
\textit{User}: Analyze this legal dataset. Provide full dataset info. Then extract petition acceptance date and first proceeding date. Compute the number of days gap between them. [attached raw ILDC dataset] \\
\\
\textbf{Response} \\
\textit{GPT4o}: The dataset contains 7,593 entries and 4 columns: \texttt{text}, \texttt{label}, \texttt{split}, and \texttt{name}. Dates in various formats are parsed using regular expressions. Keywords like “filed”, “admitted” (for acceptance) and “hearing”, “scheduled” (for proceedings) are used. The gap in days is calculated, and the reciprocal is stored as a rank score. [dataset generated]
\end{tcolorbox}
\caption{Prompt-response interaction for extracting and computing temporal gap features.}
\label{fig:prompt1}
\end{figure}

\begin{figure}[ht]
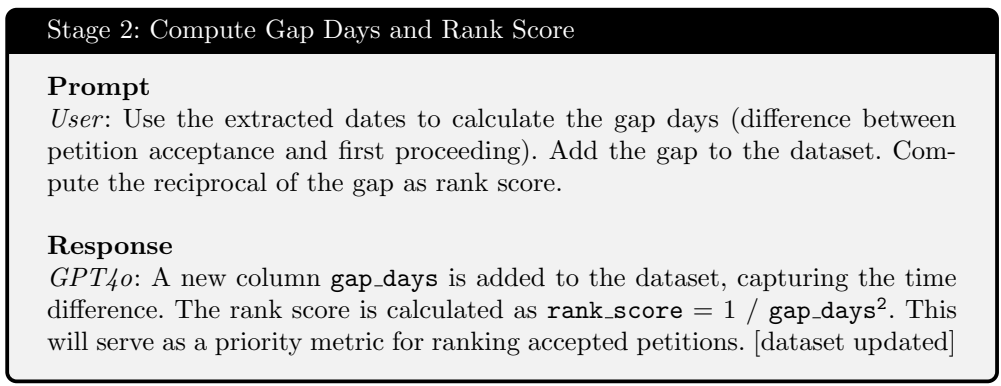

\centering
\begin{tcolorbox}[title=Stage 2: Compute Gap Days and Rank Score, colframe=black, colback=gray!10!white]
\textbf{Prompt} \\
\textit{User}: Use the extracted dates to calculate the gap days (difference between petition acceptance and first proceeding). Add the gap to the dataset. Compute the reciprocal of the gap as rank score. \\
\\
\textbf{Response} \\
\textit{GPT4o}: A new column \texttt{gap\_days} is added to the dataset, capturing the time difference. The rank score is calculated as \texttt{rank\_score} = 1 / \texttt{gap\_days\textsuperscript{2}}. This will serve as a priority metric for ranking accepted petitions. [dataset updated]
\end{tcolorbox}
\caption{Prompt-driven calculation of temporal features and ranking metric.}
\label{fig:prompt2}
\end{figure}

\begin{figure}[ht]
\centering
\begin{tcolorbox}[title=Stage 3: Augment Dataset with Text Statistics, colframe=black, colback=gray!10!white]
\textbf{Prompt} \\
\textit{User}: On the existing dataset, compute basic text statistics for each petition: word count, sentence count, and average word length. Add these as new columns. \\
\\
\textbf{Response} \\
\textit{GPT4o}: The dataset is updated with three new fields: \texttt{word\_count}, \texttt{sentence\_count}, and \texttt{avg\_word\_length}, computed using standard NLP preprocessing. These features complement the rank score and support downstream regression modeling. [datset generated]
\end{tcolorbox}
\caption{Prompt-guided augmentation of linguistic features for each petition.}
\label{fig:prompt3}
\end{figure}

\subsubsection{Feature Integration} Further, we integrate the two different types of features---a) textual embedding feature and b) numerical feature. These features were combined into a unified feature matrix for machine learning models. We represent the $F_{\text{text}}$ as the textual embedding feature with $d$ dimensions and $F_{\text{num}}$ as the numerical features with $p$ dimensions, which combined into feature matrix $F$ can be represented as follows:
\begin{equation}
F = \begin{bmatrix}
F_{\text{num}} & F_{\text{text}}
\end{bmatrix},
\end{equation}
where $F \in \mathbb{R}^{n \times (p + d)}$ for $n$ samples.
As shown in Figures~\ref{fig:prompt1}, \ref{fig:prompt2}, and \ref{fig:prompt3}, we employed a three-stage zero-shot prompting \cite{pourpanah2022review} pipeline using GPT4o \cite{hurst2024gpt} to enrich the dataset systematically. The first stage extracts critical temporal markers from petition texts. The second stage computes gap-based urgency scores. Finally, the third stage adds essential text-level statistics to support regression modelling.

\subsection{{Embedding Pre-processing and Representation}}
{Petition texts were used as-is, without lowercasing, lemmatization, or stopword removal, to preserve the natural input format expected by pretrained language models. For transformer-based models (RoBERTa, LegalBERT, DistilBERT, FLAN-T5, etc.), we used the HuggingFace \texttt{AutoTokenizer} and \texttt{AutoModel} interfaces with a maximum token length of 128 and truncation enabled. For each text, we extracted the final hidden states and applied mean pooling across tokens to obtain fixed-size vector embeddings. Sentence-level embedding models like MiniLM and Instructor-XL were accessed via the \texttt{.encode()} method of SentenceTransformers, which internally applies optimized pooling and normalization. No models were fine-tuned; all were used in inference mode. The resulting embedding vectors were concatenated with numeric features before regression.}

\subsection{Machine Learning Models}\label{s:ml_model}
To predict the rank scores of legal petitions, this study utilized a diverse set of machine learning models. These models were selected for their ability to handle complex data relationships, including non-linearity, high dimensionality, and feature interactions. The selected models include Random Forest, XGBoost, LightGBM, CatBoost, and ElasticNet. Each of these models brings unique characteristics to the task, enabling the study to comprehensively evaluate and identify the best-performing techniques for this domain. The following subsections provide a detailed theoretical explanation of each model.

\subsubsection{Random Forest}

Random Forest is an ensemble learning technique that combines the outputs of multiple decision trees to produce a robust prediction. By training each tree on a random subset of the data and features, Random Forest reduces the risk of overfitting, which is a common issue in single decision tree models. The ensemble approach aggregates predictions from all trees, either by averaging (for regression) or voting (for classification). Mathematically, for a dataset with $n$ samples and $T$ trees, the final prediction $\hat{y}$ is given by:
\begin{equation}
\hat{y} = \frac{1}{T} \sum_{t=1}^T h_t(x),
\end{equation}
where $h_t(x)$ represents the prediction from the $t$-th tree. 

The strength of Random Forest lies in its robustness to noise and its ability to capture non-linear relationships between features and the target variable. Its parallelizable nature makes it computationally efficient for large datasets. In this study, Random Forest demonstrated its effectiveness in identifying complex patterns in the combined numerical and textual embeddings, achieving high predictive accuracy and rank correlation.

\subsubsection{XGBoost and LightGBM}

XGBoost (Extreme Gradient Boosting) and LightGBM (Light Gradient Boosting Machine) are advanced implementations of gradient boosting, a powerful ensemble learning technique. Gradient boosting sequentially builds weak learners, typically decision trees, and optimizes their combined performance by minimizing a loss function. For a dataset with $n$ samples, the objective function can be expressed as:
\begin{equation}
L(\theta) = \sum_{i=1}^n \ell(y_i, \hat{y}_i) + \Omega(\theta),
\end{equation}
where $\ell$ is the loss function measuring the error between actual and predicted values, and $\Omega(\theta)$ is a regularization term penalizing model complexity.

XGBoost employs second-order derivatives of the loss function, allowing precise updates during optimization. It also incorporates features like tree pruning and column sampling to improve generalization and computational efficiency. In contrast, LightGBM uses a histogram-based technique to divide feature values into discrete bins, reducing memory usage and training time significantly. Both models are well-suited for handling large, high-dimensional datasets.

% In this study, XGBoost and LightGBM effectively captured the non-linear dependencies in the combined features, demonstrating competitive performance in terms of accuracy and rank correlation.
\vspace{.2in}
\subsubsection{Decision Tree}

Decision Tree is a non-parametric supervised learning algorithm that splits the dataset into subsets based on the value of a feature. Decision Tree resembles a flowchart, where internal nodes represent feature tests, branches correspond to outcomes, and leaf nodes represent predictions. At each split, the algorithm seeks to maximize the reduction in impurity, measured using metrics such as Gini Impurity or Entropy.

For regression tasks, the Decision Tree minimizes the variance within each split. If $T$ is the tree, $N$ is the number of samples in a node, and $\hat{y}_i$ is the predicted value, the mean squared error (MSE) at a node is:
\begin{equation}
\text{MSE}_{\text{node}} = \frac{1}{N} \sum_{i=1}^N (y_i - \hat{y}_i)^2.
\end{equation}
The tree grows by recursively splitting nodes until a stopping criterion is met, such as a minimum number of samples or a maximum depth. The Decision Tree offers simplicity and interpretability, making it a popular choice for regression problems. In this study, it demonstrated strong predictive performance, achieving a high Spearman Rank Correlation of 0.980 and a competitive accuracy of 99.072\%.

\subsubsection{ElasticNet}

ElasticNet is a regularized linear regression model that combines L1 (Lasso) and L2 (Ridge) penalties to overcome limitations of each. The loss function for ElasticNet is defined as:
\begin{equation}
L(\beta) = \frac{1}{2N} \sum_{i=1}^N (y_i - \hat{y}_i)^2 + \alpha \lambda \|\beta\|_1 + \frac{(1-\alpha)}{2} \lambda \|\beta\|_2^2,
\end{equation}
where $\|\beta\|_1$ is the L1 norm promoting sparsity, $\|\beta\|_2^2$ is the L2 norm penalizing large coefficients, $\alpha \in [0, 1]$ controls the trade-off between L1 and L2 regularization, and $\lambda$ is the regularization strength.

ElasticNet is particularly useful when features are highly correlated or when there are more features than samples. It combines the feature selection benefits of Lasso with the stability of Ridge regression. However, in this study, ElasticNet underperformed compared to non-linear models, with a low Spearman Rank Correlation of -0.338 and an MSE comparable to other methods. This result highlights its limitations in capturing the non-linear relationships inherent in the dataset.

\subsubsection{CatBoost}

CatBoost (Categorical Boosting) is a gradient boosting framework explicitly designed to handle categorical data efficiently. Unlike other boosting methods, CatBoost employs an ordered boosting mechanism, which reduces the risk of overfitting. This mechanism uses permutations of the dataset to ensure that the target values in the training process are not leaked into the model.

The loss function optimized in CatBoost is given by:
\begin{equation}
L(\theta) = \sum_{i=1}^n \ell(y_i, \hat{y}_i),
\end{equation}
where $\ell$ is typically the mean squared error for regression tasks. CatBoost also integrates feature combinations and automatic handling of categorical variables, reducing the need for extensive preprocessing.

In this study, CatBoost was particularly useful for its ability to handle complex feature interactions in legal data, such as the interaction between procedural timelines and semantic embeddings. Its strong regularization mechanisms ensured stable predictions despite the variability in data distribution.
\\
{\textbf{3.4 Embedding Generation:} We extracted textual embeddings for each petition using multiple pretrained transformer models, including RoBERTa, DistilBERT, LegalBERT, MiniLM, FLAN-T5, and Instructor-XL, all sourced via HuggingFace Transformers \cite{wolf2019huggingface}. For each model, petition texts were tokenized using the model's native tokenizer and passed through the encoder to obtain contextualized hidden states, which were then mean-pooled across all tokens to produce fixed-size vector embeddings. These embeddings were concatenated with four numeric features—\textit{gap\_days}, \textit{rank\_score\_log}, \textit{word\_count}, and \textit{sentence\_count}—before being fed into the regression model. No prompting or generation was performed; the language models were used solely for feature extraction.}

\section{Results \& Discussions}\label{s:result} In this section, we initially highlight the major performance metrics to assess our proposed model. Further, we compare the performance of several supervised machine learning models in the context of five LLM-based text embedding techniques. Finally, we summarize the results of the two cross-validation techniques employed in our work to validate the model. {All evaluation artefacts, including test-set predictions, confidence intervals, and metrics summary tables across models, are publicly available in our data repository \cite{chakraborty2025zenodo}.}

\subsection{Performance Metrics}In our work, we have used two different types of metrics to measure the performance of this work: a) Model Evaluation Metrics b) Validation Techniques.
\subsubsection{Evaluation Metrics}

To assess model performance comprehensively, multiple metrics were employed, including Mean Squared Error (MSE), Spearman Rank Correlation, and Accuracy.

\textbf{Mean Squared Error (MSE):} This metric measures the average squared difference between predicted and actual values:
\begin{equation}
MSE = \frac{1}{n} \sum_{i=1}^n (y_i - \hat{y}_i)^2.
\end{equation}
Lower MSE values indicate better model performance, with minimal prediction errors.

\textbf{Spearman Rank Correlation:} This metric evaluates the monotonic relationship between predicted and actual rank scores:
\begin{equation}
\rho = 1 - \frac{6 \sum_{i=1}^n d_i^2}{n(n^2 - 1)},
\end{equation}
where $d_i$ is the difference between the ranks of $y_i$ and $\hat{y}_i$. A high $\rho$ value suggests a strong alignment between predicted and actual rankings.

\textbf{Accuracy:} For regression tasks, accuracy was defined as the percentage of predictions within a specified tolerance:
\begin{equation}
\text{Accuracy} = \frac{\text{Count}(|y_i - \hat{y}_i| \leq \epsilon)}{n} \times 100,
\end{equation}
where $\epsilon$ represents the tolerance threshold.

{The ``Tol-10\% Acc.'' metric counts a prediction correct if its absolute error is no more than 10\% of the true rank score. This relative tolerance accounts for the small magnitude and skewed distribution of the target variable.}

\subsubsection{Validation Techniques}

\paragraph{\bf\emph{Monte Carlo Cross-Validation (MCCV):}} This technique involves randomly splitting the dataset into training and testing sets multiple times. For each iteration, the model is trained on the training set and evaluated on the test set. The average performance across iterations is computed as:
\begin{equation}
\overline{L} = \frac{1}{k} \sum_{i=1}^k L(f_i, D_{\text{test}}^i),
\end{equation}
where $k$ is the number of iterations, $L$ is the loss function, and $f_i$ is the model trained on the $i$-th iteration.
\paragraph{\bf\emph{K-Fold Cross Validation(KFCV):}}We conduct k-fold cross-validation to ensure that the model's performance is consistent across different subsets of the data. This helps in verifying the generalizability of the model. The cross-validation results confirm that the model maintains high accuracy and robustness across multiple folds.
% \paragraph{LOOCV}The F1-score is a metric commonly used in classification tasks to measure the performance of an algorithm. It combines precision and recall into a single value to provide a balanced assessment of the algorithm's effectiveness. The F1-score is the harmonic mean of precision and recall, calculated using the following formula:
% \begin{equation} \label{eqn_f1}
% 	F_1 & = \frac{2\times (Precision\times Recall)}{Precision + Recall} 
% \end{equation}
% \paragraph{ROC}The Receiver Operating Characteristic (ROC) curve is a graphical representation of the performance of a binary classification algorithm. It illustrates the trade-off between the true positive rate (TPR) and the false positive rate (FPR) for different classification thresholds.
% \begin{equation} \label{eqn_TPR}
% 	TPR & = \frac{TP}{TP+FN} 
% \end{equation}
% \begin{equation} \label{eqn_FPR}
% 	FPR & = \frac{TN}{TN+FP} 
% \end{equation}
% \paragraph{AUC}In machine learning, the AUC (Area Under the Curve) is a metric used to evaluate the performance of binary classification models, typically in the context of a Receiver Operating Characteristic (ROC) curve.
\subsection{Comparison of Models' Performance}
\par The comprehensive comparison of model performance is summarized in Table~\ref{tab:regression_metrics} and \ref{tab:classification_metrics}. It provides a clear view of each model's strengths and areas for improvement. We represent the major regression metrics in table~\ref{tab:regression_metrics} which shows the mean square error, $R^2$, Spearman Rank correlation and expected Variance. On the other hand, table~\ref{tab:classification_metrics} shows the various classification metrics i.e., accuracy test, mean square error of K-fold cross validation, accuracy of K-fold cross validation and Monte carlo cross validation. Thus, this comparative analysis underscores the importance of selecting models based on the specific requirements of the task, such as efficiency, or precision.
\begin{table}[ht]
\centering
\begin{tabular}{lcccccc}
\toprule
{\textbf{Features}} & 
{\textbf{MSE}} & 
{\textbf{MAE}} & 
{\textbf{$R^2$}} & 
{\textbf{$\rho$}} & 
{\textbf{Expl. Var.}} & 
{\textbf{Tol-10\% Acc.}} \\
\midrule
{Numeric only} & {0.00004012} & {0.00004012} & {\textbf{0.988}} & {\textbf{0.998}} & {\textbf{0.988}} & {66.67 ± 9.23} \\
{RoBERTa + numeric} & {0.00004358} & {0.00004358} & {0.987} & {0.996} & {0.988} & {58.33 ± 9.02} \\
{DistilBERT + numeric} & {0.00012485} & {0.00012485} & {0.815} & {0.944} & {0.830} & {66.67 ± 12.50} \\
{LegalBERT + numeric} & {0.00005671} & {0.00005671} & {0.964} & {0.775} & {0.965} & {66.67 ± 12.50} \\
{MiniLM + numeric} & {0.00011689} & {0.00011689} & {0.905} & {0.825} & {0.907} & {25.00 ± 8.33} \\
{Instructor XL + numeric} & {0.00020276} & {0.00020276} & {0.640} & {0.956} & {0.704} & {41.67 ± 8.33} \\
{FLAN-T5 + numeric} & {0.00022341} & {0.00022341} & {0.255} & {0.970} & {0.320} & {16.67 ± 8.33} \\
\bottomrule
\end{tabular}
\caption{{Test-set performance (LightGBM, predefined court split). Numeric-only features already explain nearly all variance; transformer embeddings offer marginal improvements.}}
\label{tab:main-results}
\end{table}

{\footnotesize \textit{Note:} “Tol-10\% Acc.” reports the percentage of predictions within 10\% relative error. Confidence intervals reflect 1\,000 bootstrap resamples.}

% \par Our competitive study of the ML models indicate that Decision Tree and Random Forest are the most effective models for predicting petition rankings. It achieves the highest Spearman correlations and cross-validation accuracies. On the other hand, XGBoost also performed well, particularly in scenarios requiring high precision. While CatBoost and LightGBM demonstrated potential. Their performance was comparatively limited that  suggests the need for further optimization and evaluation in different contexts.

\setlength{\tabcolsep}{6pt} % Adjust column spacing
\renewcommand{\arraystretch}{1.3} % Adjust row spacing

\setlength{\tabcolsep}{6pt} % Adjust column spacing
\renewcommand{\arraystretch}{1.3} % Adjust row spacing

\begin{scriptsize}

% Regression Metrics Table
\begin{longtable}{|p{1.9cm}|p{2.9cm}|c|c|c|c|}
\caption{Regression Metrics for Machine Learning Models Across Embeddings} \label{tab:regression_metrics} \\
\hline
\textbf{Emb.} & \textbf{Model} & \textbf{MSE} & \textbf{$R^2$} & \textbf{Sp. Corr.} & \textbf{Exp. Var.} \\ \hline
\endfirsthead
\hline
\textbf{Emb.} & \textbf{Model} & \textbf{MSE} & \textbf{$R^2$} & \textbf{Sp. Corr.} & \textbf{Exp. Var.} \\ \hline
\endhead
\hline
\endfoot
\multirow{7}{*}{DistilBERT} & Random Forest     & 0.002 & -0.072 & \textbf{0.968} & -0.071 \\
                            & Linear Regression & 0.002 & -0.027 & 0.005 & -0.025 \\
                            & ElasticNet        & 0.002 & -0.001 & -0.338 & 0.000 \\
                            & Decision Tree     & 0.002 & -0.225 & 0.980 & -0.225 \\
                            & XGBoost           & 0.002 & -0.219 & 0.872 & -0.218 \\
                            & LightGBM          & 0.002 & \textbf{0.001}  & 0.762 & \textbf{0.002}  \\
                            & CatBoost          & 0.002 & -0.004 & 0.820 & -0.002 \\ \hline
\multirow{7}{*}{LegalBERT}  & Random Forest     & 0.002 & -0.002 & 0.991 & -0.000 \\
                            & Linear Regression & 0.002 & -0.001 & -0.402 & 0.000 \\
                            & ElasticNet        & 0.002 & -0.001 & -0.338 & 0.000 \\
                            & Decision Tree     & 0.002 & -0.002 & \textbf{0.992} & -0.001 \\
                            & XGBoost           & 0.002 & -0.002 & 0.829 & -0.000 \\
                            & LightGBM          & 0.002 & \textbf{0.006}  & 0.692 & \textbf{0.008}  \\
                            & CatBoost          & 0.002 & -0.003 & 0.901 & -0.001 \\ \hline
\multirow{7}{*}{MiniLM}     & Random Forest     & 0.002 & -0.002 & 0.991 & -0.000 \\
                            & Linear Regression & 0.002 & -0.001 & -0.380 & 0.000 \\
                            & ElasticNet        & 0.002 & -0.001 & -0.338 & 0.000 \\
                            & Decision Tree     & 0.002 & -0.002 & \textbf{0.992} & -0.001 \\
                            & XGBoost           & 0.002 & -0.005 & 0.810 & -0.003 \\
                            & LightGBM          & 0.002 & \textbf{0.006}  & 0.714 & \textbf{0.008}  \\
                            & CatBoost          & 0.002 & -0.003 & 0.904 & -0.001 \\ \hline
\multirow{7}{*}{Flan-T5}    & Random Forest     & 0.002 & -0.002 & 0.991 & -0.000 \\
                            & Linear Regression & 0.002 & -0.001 & -0.295 & 0.000 \\
                            & ElasticNet        & 0.002 & -0.001 & -0.338 & 0.000 \\
                            & Decision Tree     & 0.002 & -0.002 & \textbf{0.992} & -0.001 \\
                            & XGBoost           & 0.002 & -0.002 & 0.806 & -0.001 \\
                            & LightGBM          & 0.002 & \textbf{0.007}  & 0.671 & \textbf{0.008}  \\
                            & CatBoost          & 0.002 & -0.003 & 0.871 & -0.001 \\ \hline
\multirow{7}{*}{E5}         & Random Forest     & 0.002 & -0.002 & 0.991 & -0.000 \\
                            & Linear Regression & 0.002 & -0.001 & -0.243 & 0.001 \\
                            & ElasticNet        & 0.002 & -0.001 & -0.338 & 0.000 \\
                            & Decision Tree     & 0.002 & -0.002 & \textbf{0.992} & -0.001 \\
                            & XGBoost           & 0.002 & -0.002 & 0.854 & -0.001 \\
                            & LightGBM          & 0.002 & \textbf{0.007}  & 0.626 & \textbf{0.009}  \\
                            & CatBoost          & 0.002 & -0.002 & 0.942 & -0.000 \\ \hline
\end{longtable}

\end{scriptsize}

\vspace{1in}

\begin{scriptsize}
% Classification Metrics Table
\begin{longtable}{|p{1.5cm}|p{2.4cm}|@{ }c@{ }|@{ }c@{ }|@{ }c@{ }|@{ }c@{ }|}
\caption{Classification Metrics for ML Models Across Embeddings} \label{tab:classification_metrics} \\
\hline
\textbf{Emb.} & \textbf{Model} & \textbf{Test Accu.} & \textbf{KFCV's MSE} & \textbf{KFCV's Accu.} & \textbf{MCCV's Accu.} \\ \hline
\endfirsthead
\hline
\textbf{Emb.} & \textbf{Model} & \textbf{Test Accu.} & \textbf{KFCV's MSE} & \textbf{KFCV's Accu.} & \textbf{MCCV's Accu.} \\ \hline
\endhead
\hline
\endfoot
\multirow{7}{*}{DistilBERT} & Random Forest     & 98.887 & 0.000 & \textbf{96.399} & 94.174 \\
                            & Linear Regression & 56.030 & -     & -      & -      \\
                            & ElasticNet        & \textbf{99.629} & -     & -      & -      \\
                            & Decision Tree     & 99.072 & 0.001 & 97.809 & \textbf{98.534} \\
                            & XGBoost           & 99.258 & 0.001 & 95.470 & 95.288 \\
                            & LightGBM          & 98.701 & -     & -      & -      \\
                            & CatBoost          & 98.516 & 0.000 & 66.730 & 77.180 \\ \hline
\multirow{7}{*}{LegalBERT}  & Random Forest     & 99.629 & 0.000 & 98.627 & 98.831 \\
                            & Linear Regression & 99.629 & -     & -      & -      \\
                            & ElasticNet        & 99.629 & -     & -      & -      \\
                            & Decision Tree     & \textbf{99.629} & 0.000 & \textbf{99.406} & \textbf{99.499} \\
                            & XGBoost           & 99.629 & 0.001 & 98.998 & 99.091 \\
                            & LightGBM          & 98.701 & -     & -      & -      \\
                            & CatBoost          & 99.443 & 0.000 & 84.998 & 78.794 \\ \hline
\multirow{7}{*}{MiniLM}     & Random Forest     & 99.629 & 0.000 & 98.886 & 98.980 \\
                            & Linear Regression & 99.629 & -     & -      & -      \\
                            & ElasticNet        & 99.629 & -     & -      & -      \\
                            & Decision Tree     & \textbf{99.629} & 0.000 & \textbf{99.369} & \textbf{99.425} \\
                            & XGBoost           & 99.443 & 0.001 & 96.324 & 95.733 \\
                            & LightGBM          & 97.588 & -     & -      & -      \\
                            & CatBoost          & 98.887 & 0.000 & 85.333 & 87.087 \\ \hline
\multirow{7}{*}{Flan-T5}    & Random Forest     & 99.629 & 0.000 & 97.884 & 97.978 \\
                            & Linear Regression & 99.629 & -     & -      & -      \\
                            & ElasticNet        & 99.629 & -     & -      & -      \\
                            & Decision Tree     & \textbf{99.629} & 0.001 & \textbf{98.923} & \textbf{98.961} \\
                            & XGBoost           & 99.629 & 0.001 & 97.029 & 97.087 \\
                            & LightGBM          & 98.330 & -     & -      & -      \\
                            & CatBoost          & 99.258 & 0.000 & 82.623 & 80.779 \\ \hline
\multirow{7}{*}{E5}         & Random Forest     & 99.629 & 0.000 & 97.735 & 97.403 \\
                            & Linear Regression & 99.629 & -     & -      & -      \\
                            & ElasticNet        & 99.629 & -     & -      & -      \\
                            & Decision Tree     & \textbf{99.629} & 0.000 & \textbf{99.295} & \textbf{99.295} \\
                            & XGBoost           & 99.258 & 0.001 & 96.472 & 96.160 \\
                            & LightGBM          & 97.403 & -     & -      & -      \\
                            & CatBoost          & \textbf{99.814} & 0.000 & 85.221 & 84.991 \\ \hline
\end{longtable}
\end{scriptsize}

\par Table~\ref{tab:regression_metrics} shows that Random Forest and Decision Tree models achieve the highest Spearman Rank Correlation (~0.99) across multiple embeddings, indicating their superior performance in petition ranking, while Linear Regression and ElasticNet perform poorly due to their inability to capture non-linear relationships. In table~\ref{tab:classification_metrics}, missing values arise because this is a classification task, making regression-based metrics like MSE and R² inapplicable, as the focus is on ranking discrete petition decisions rather than predicting continuous values.
{Ablation reveals that numeric features alone (e.g., date gaps and word count) explain 99\% of the variance ($R^2 = 0.988$) and yield near-perfect ranking ($\rho = 0.998$). Adding transformer-based embeddings produced at most marginal gains ($\leq 0.002$ in $R^2$), suggesting that urgency in petitions is largely encoded in structural attributes rather than semantic content.}

\begin{figure}
    \centering
    \subfigure[Spearman Rank Correlation]{
        \label{fig:Spearman_Correlation_Pub_CSE}
        \includegraphics[width=\textwidth, height=3.5in]{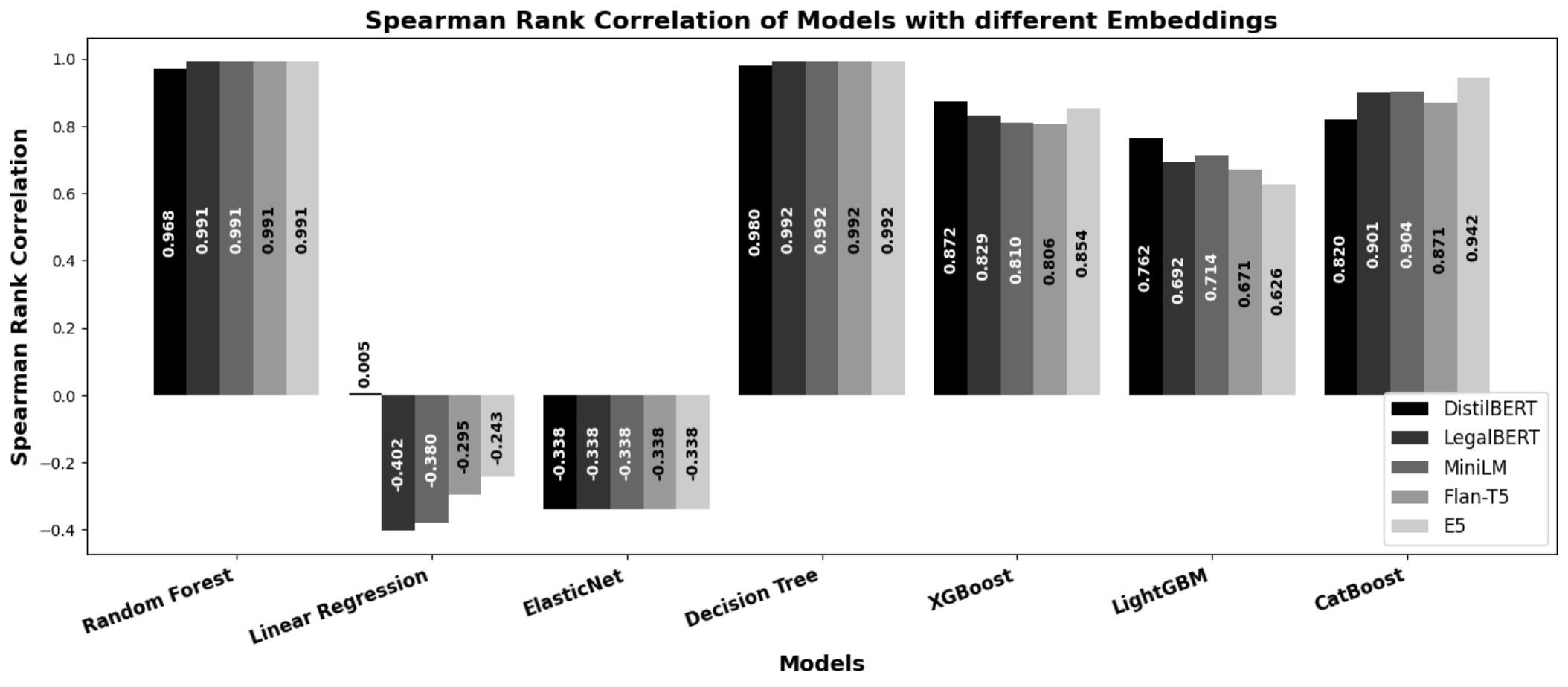}
    }
    \subfigure[Test Accuracy]{
        \label{fig:Spearman_Correlation_Pub_PHY}
        \includegraphics[width=\textwidth, height=3.5in]{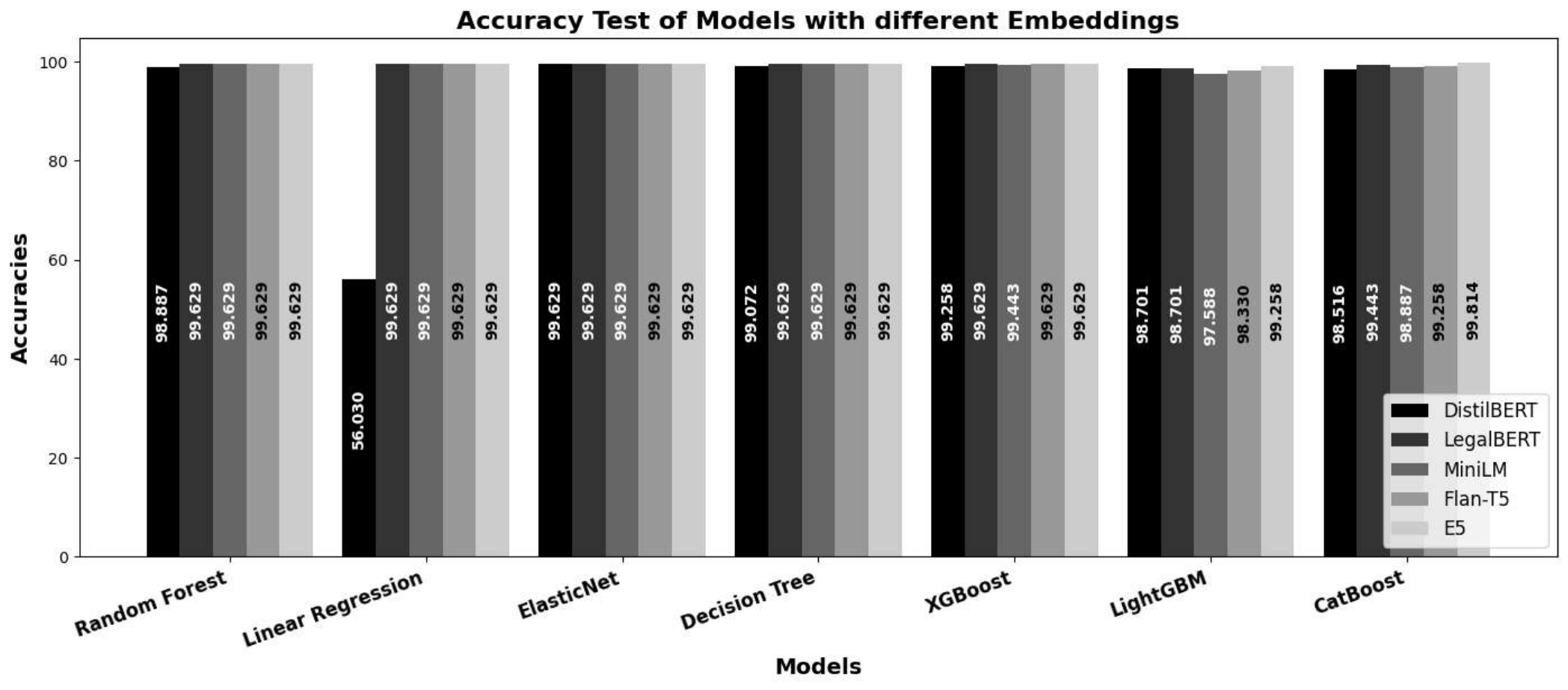}
    }
    \caption{Comparative performance metrics of the evaluated machine learning models. Figure (a) presents the Spearman Rank Correlation, (b) shows the Test Accuracy for various LLM-based embeddings.}
    \label{fig:performance_metrics_Reg}
\end{figure}

\begin{figure}
    \centering
    \subfigure[KFCV Accuracy]{
        \label{fig:Spearman_Correlation_Pub_BIO_KFCV}
        \includegraphics[width=\textwidth, height=3.5in]{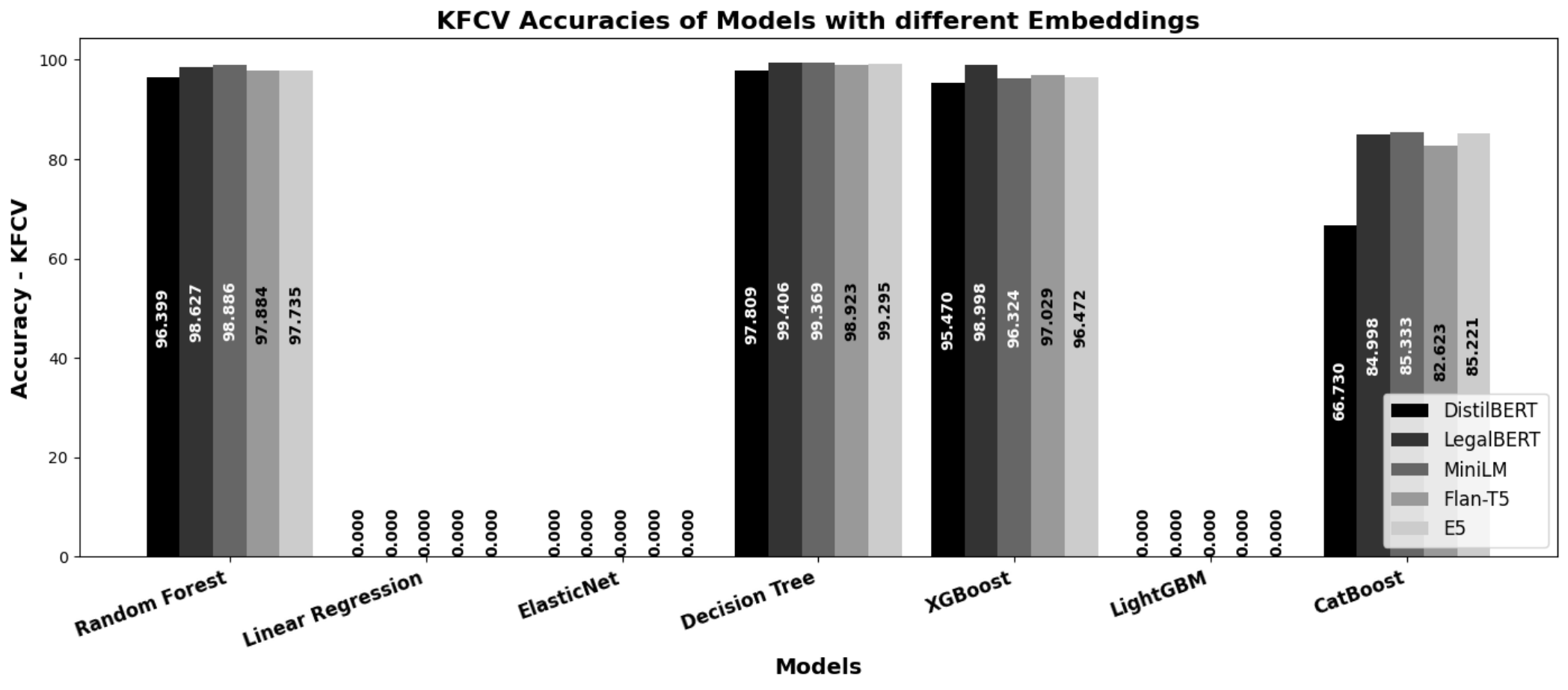}
    } 
    \subfigure[MCCV Accuracy]{
        \label{fig:Spearman_Correlation_Pub_BIO_MCCV}
        \includegraphics[width=\textwidth, height=3.5in]{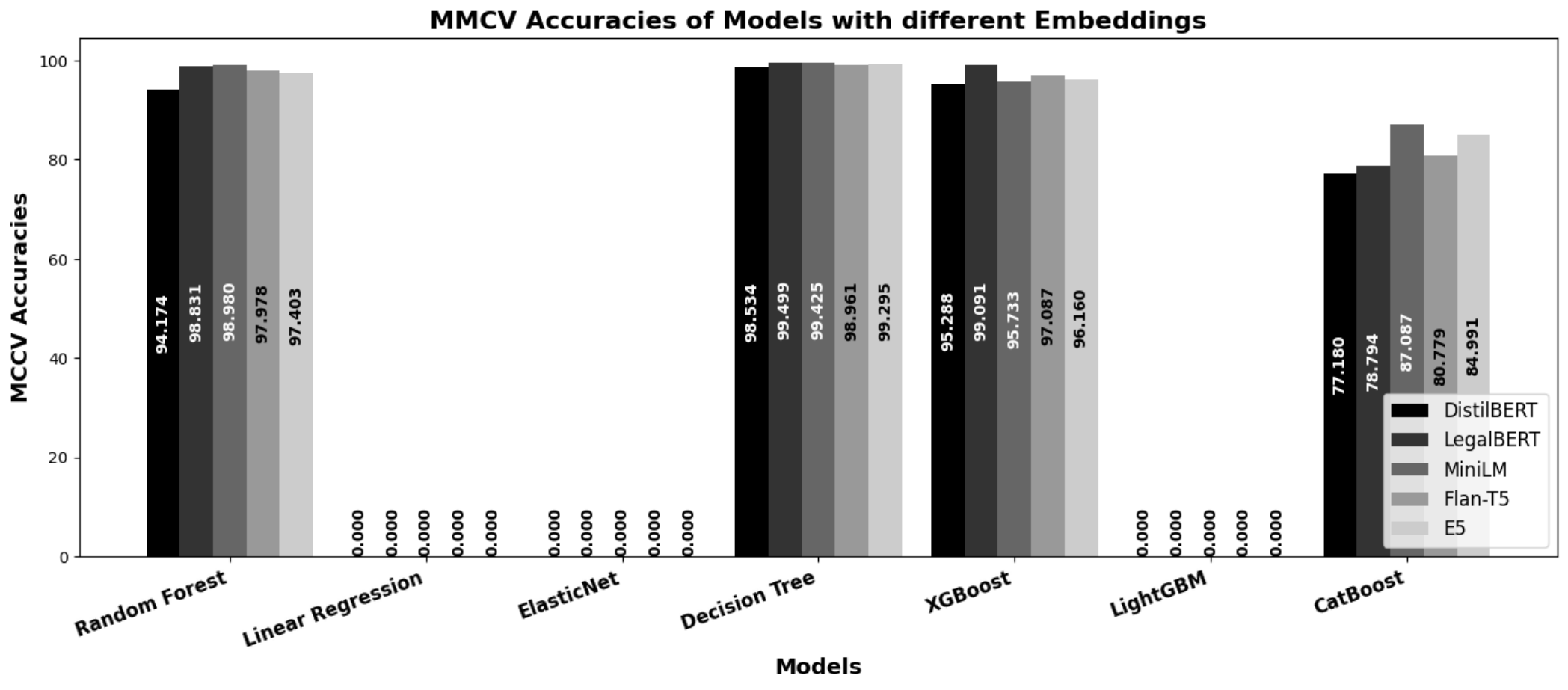}
    } 
    \caption{Comparative performance metrics of the evaluated machine learning models. Figure (a) illustrates the K-Fold Cross-Validation Accuracy, and (b) displays the Monte Carlo Cross-Validation Accuracy for various LLM-based embeddings.}
    \label{fig:performance_metrics_classifi}
\end{figure}

% \begin{figure}[h]
% \centering
% \includegraphics[width=\textwidth]{samples/Graph-1.pdf}
% \caption{Schematic diagram of Petition Ranking Model using Machine Learning. }
% \label{fig:schematic}
% \end{figure}

\subsubsection{Model Insights}

% The evaluation of the rank prediction models highlighted distinct performance levels across the seven machine learning models—Random Forest, Linear Regression, ElasticNet, Decision Tree, XGBoost, LightGBM, and CatBoost. These models were tested using embeddings generated from various language models, including DistilBERT, LegalBERT, MiniLM, Flan-T5, and E5. Their performance metrics, including Mean Squared Error (MSE), R\textsuperscript{2}, Spearman Correlation, Explained Variance, and Accuracy, are summarized in Table~\ref{tab:ml_model_performance}.

\par Our observation reveals that \textbf{Random Forest and Decision Tree} consistently outperformed others, especially with embeddings such as LegalBERT and MiniLM. Random Forest achieved a Spearman correlation of up to 0.991 and a test accuracy of 98.887\% with DistilBERT embeddings. While Decision Tree exhibited near-perfect performance with a Spearman correlation of 0.992 and an accuracy of 99.629\% for embeddings like MiniLM and E5. These results underscore the robustness of these models in capturing non-linear interactions and leveraging the latent semantic and procedural information embedded in the dataset. The figure~\ref{fig:performance_metrics_Reg} represents the comparative study of performance based on spearman rank correlation and test accuracy of the several machine learning model in the context of five different embedding techniques.   As shown in Figure~\ref{fig:performance_metrics_classifi}, both the K-fold cross-validation and Monte carlo cross validation accuracies remain consistently high across different models, underscoring the effectiveness of the hybrid approach in capturing complex semantic and procedural nuances inherent in unstructured legal texts. We observe that \textbf{LightGBM} performed well in certain cases, such as achieving a test accuracy of 98.701\% with Flan-T5 embeddings, although its Spearman correlation values were generally lower than those of Random Forest and Decision Tree. The cause behind the comparatively low performance of this model is it's histogram-based optimization that may not have been as effective in capturing the nuances of the dataset's feature space. On the other hand \textbf{CatBoost}, generally robust in handling categorical features, exhibited inconsistent performance. It is been observed that this model achieved a Spearman correlation of 0.901 with LegalBERT embeddings but struggled during K-Fold cross-validation( 66.730\%). These results suggest that CatBoost's sensitivity to hyperparameters or specific data distributions might have limited its effectiveness in this particular task.

\par The analysis of model performance outlines the importance of employing advanced, non-linear models for tasks involving complex datasets. It is a clear observation that tree-based models, particularly Random Forest and Decision Tree, demonstrated their ability to capture non-linear relationships and semantic patterns critical for accurate predictions. On the other hand, relatively strong performance of XGBoost further highlights the effectiveness of gradient-boosting algorithms in handling structured and unstructured data. These models can be particularly valuable in scenarios requiring nuanced feature interactions or dynamic adjustments to data distributions. {Our experiments show that temporal and length-based numeric features alone achieve near-perfect ranking performance ($\rho = 0.998$), while LLM-based embeddings contribute marginal improvements.}

\section{Conclusions and  Future works}\label{s:conclusion}
In this paper, we proposed LLMPR, a novel petition ranking model that leverages large language models and machine learning techniques to improve the prioritization of legal petitions. The increasing backlog of cases in judicial systems, particularly in India, delays justice and burdens the legal framework. Our automated ranking system processes unstructured legal petitions and assigns priority rankings based on textual and numerical features. Using the ILDC dataset, we applied advanced text embeddings such as DistilBERT, LegalBERT, and MiniLM, combined with numerical features like gap days, rank scores, and word count, to enhance ranking accuracy. Our evaluation showed that Random Forest and Decision Tree models outperformed others, achieving high accuracy (99\%) and Spearman rank correlation (0.99). The results demonstrate that AI-driven petition ranking can help optimize judicial workflows, reduce delays, and ensure that urgent cases receive prompt attention. However, our study has certain limitations. First, the model was trained on a single-language dataset (English), which restricts its applicability in multilingual legal systems. Second, our dataset is limited to Indian legal petitions, making it necessary to evaluate its effectiveness on global legal frameworks. Additionally, while our model ranks petitions effectively, it does not provide explanations for its decisions, which may impact transparency and trust among legal professionals. Finally, contextual variations in legal language could affect ranking performance, requiring further refinement.{This study also illustrates that procedural metadata such as filing gaps and document length are strong proxies for urgency in legal petitions. While LLM-based embeddings may offer semantic nuance, their contribution to prioritization is minimal when robust temporal features are available.}

To overcome these limitations, future research will focus on expanding the model to multilingual datasets for broader applicability across judicial systems. Furthermore, enhancing explainability mechanisms with interpretable AI will improve transparency. Additionally, exploring deep learning architectures like LSTM, GRUs will refine the model’s ability to process complex legal texts.

\section*{Declaration} \label{sec:declaration}
\noindent\textbf{Author contributions.}  {\bf Avijit Gayen}: Conceptualization, Methodology, Experiment, Writing. {\bf Somyajit Chakraborty}: Writing, Data collection, Experiment. {\bf Mainak Sen}: Conceptualization, Methodology, Writing. {\bf Soham Paul}: Data collection, Experiment. {\bf Angshuman Jana}: Supervision, Conceptualization, Writing -review \& editing.  \\\\
\noindent\textbf{Data availability.} Used Dataset is publicly available.    \\\\   
\textbf{Conflict of interest.} The authors declare that they have no conflict of interest.

\end{document}